\documentclass[letterpaper, 10 pt, conference]{ieeeconf}

\IEEEoverridecommandlockouts
\usepackage{cite}
\usepackage{times}
\usepackage[pdftex]{graphicx}
\graphicspath{{../Figures}}

\DeclareGraphicsExtensions{.pdf,.jpeg,.png,.jpg}
\usepackage{caption}
\usepackage[normalem]{ulem}
\usepackage{subcaption}
\usepackage{epstopdf}
\usepackage[cmex10]{amsmath}
\usepackage{amssymb}
\usepackage{amsfonts}
\usepackage{mathrsfs}
\usepackage{amsbsy}
\usepackage{mathabx}
\usepackage{algorithmic}
\usepackage{array}
\usepackage{mdwmath}
\usepackage{mdwtab}
\usepackage{stfloats}
\usepackage{fancybox,color}
\usepackage[lined,commentsnumbered]{algorithm2e}
\usepackage{url}
\usepackage[utf8]{inputenc}
\usepackage[nodisplayskipstretch]{setspace}
\usepackage{hyperref}
\usepackage{ulem}
\hypersetup{
    colorlinks=true,
    linkcolor=blue,
    filecolor=blue,      
    urlcolor=blue,
}
\usepackage{lipsum}

\def\eg{e.g., }

\newcommand\figref{Fig.~\ref}

\newcommand\eqnref{Eq.~\eqref}

\newcommand{\rulesep}{\unskip\ \vrule width .8pt\ }

\usepackage{ulem}

\title{\LARGE \bf
Bayesian Optimization-based Nonlinear Adaptive PID Controller \\ 
 Design for Robust Mobile Manipulation
}

\author{Hadi Hajieghrary$^{1}$, Marc Peter Deisenroth$^2$, and Yasemin Bekiroglu$^{1,2}$
            \thanks{$^{1}$Chalmers University of Technology, Department of Electrical Engineering, Automatic Control research group, Sweden. $^{2}$University College London, Department of Computer Science, Centre for Artificial Intelligence, U.K. This work was supported by Chalmers AI Research Center (CHAIR) and Chalmers Gender Initiative for Excellence (Genie), and the project AIMCoR - AI-enhanced Mobile Manipulation Robot for Core Industrial Applications.
            Email:{~\tt\small hadiha@chalmers.se}}
}

\begin{document}

\maketitle
\thispagestyle{empty}
\pagestyle{empty}

\begin{abstract}
In this paper, we propose to use a nonlinear adaptive PID controller to regulate the joint variables of a mobile manipulator. The motion of the mobile base forces undue disturbances on the joint controllers of the manipulator. In designing a conventional PID controller, one should make a trade-off between the performance and agility of the closed-loop system and its stability margins. The proposed nonlinear adaptive PID controller provides a mechanism to relax the need for such a compromise by adapting the gains according to the magnitude of the error without expert tuning. Therefore, we can achieve agile performance for the system while seeing damped overshoot in the output and track the reference as close as possible, even in the presence of external disturbances and uncertainties in the modeling of the system. We have employed a Bayesian optimization approach to choose the parameters of a nonlinear adaptive PID controller to achieve the best performance in tracking the reference input and rejecting disturbances. The results demonstrate that a well-designed nonlinear adaptive PID controller can effectively regulate a mobile manipulator's joint variables while carrying an unspecified heavy load and an abrupt base movement occurs.
\end{abstract}








\section{Introduction}





Mobile manipulators have long been the sought-after technology to automate material handling in industrial settings. They are expected to have a central role in Industry 4.0 to revolutionize the workshops, assembly lines, warehouses, and construction sites \cite{Hamner2010, Domel2017, Brosque2020, Stibinger2021}. In recent years, mobile manipulators have appeared outside industrial settings in our daily lives as service robots \cite{ Gorjup2021}. 

Both mobile robots and robotic manipulators enjoy very mature and robust technologies; nevertheless, combining the two systems brings up new opportunities and challenges: opportunities to use the combined facilities of the mobile base and robotic manipulator to improve the performance of the overall system, increase its capability, and broaden its reach; and challenges to avoid the adverse effect that coupling the two subsystems can have on each other's stability and performance \cite{Yao2010, Bostelman2016, Seo2018}.   

\begin{figure}[t]
\centering
    \includegraphics[width=0.9\linewidth]{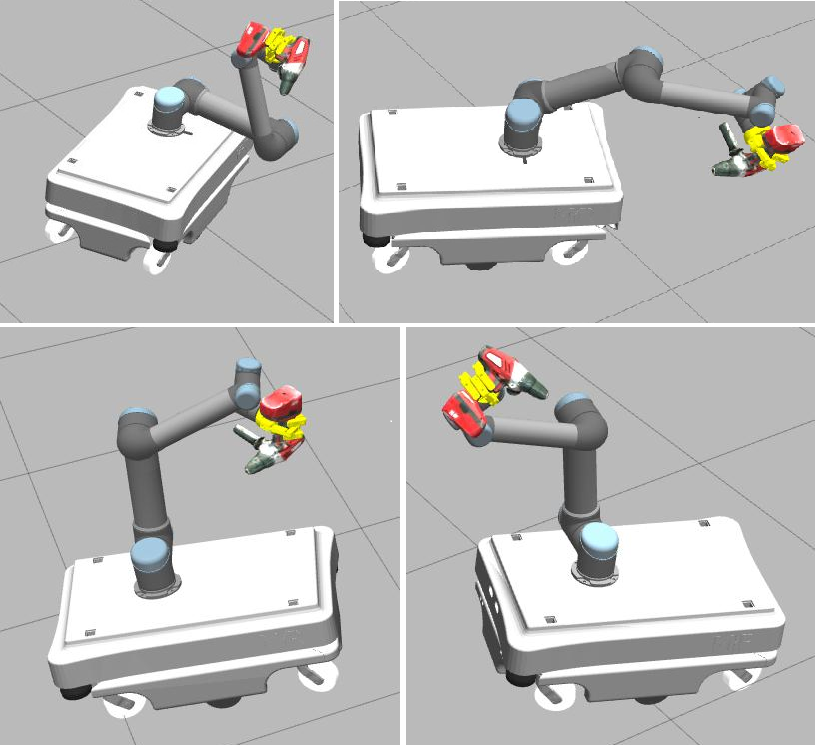}
    \caption{Gazebo simulation of grab-and-go scenario. The mobile manipulator consists of a MiR100 mobile base on which a UR5 manipulator is mounted, and a three-finger Barrett hand BH282 is attached as the end effector. The simulation scenario is to grab a heavy power drill (3.5kg), and maneuver it while the mobile base is moving around.}
    \label{simulation scenes}
     \vspace{-5mm}
\end{figure}

A mobile base allows the robot to perceive the scene from different angles and acquire a better temporal understanding \cite{Faulhammer2015, Stibinger2021}. The added degrees of freedom increase the robot's dexterity and enable new capabilities for task and motion planning \cite{Berntorp2012, Osman2020, Wang2020}. However, maneuvering the mobile base may put an undue burden on joint controllers of the manipulator, depending on the configuration of the manipulator and the bearing and acceleration of the base at the time, see \figref{simulation scenes}. Estimating the bounds of this disturbance is complex, and designing controllers to satisfy the requirements of robust performance against the adverse effects of this uncertainty is challenging. We may identify dynamic and kinematic stability regions beforehand and recommend operating the system within these boundaries \cite{Furuno2003, Moosavian2007, Bostelman2017}. However, conservative measures like these may substantially limit the versatility of a mobile manipulator since any planning for its action should navigate through these added limitations. This fact has warranted a substantial body of research to design the mechanics of a mobile manipulator to exhaust all the possibilities to increase the breadth of the stability region \cite{Gardner2000, Padois2007, Sandakalum2022}.

A robust controller has to be designed to deliver the best performance in the presence of structured and unstructured model uncertainties and external disturbances, as long as a measure of these uncertainties remains within the boundaries known when the controller is designed \cite{Sage2010, Ferrara2015, Seo2018}. However, robust controllers often give up too much of the performance to guarantee the stability of the overall system. In the case of a manipulator, the primary sources of uncertainty are various amounts of load the robot may handle during its operation and the changing configuration of the robot as it moves. The uncertainty of the load can be parameterized and modeled as structured uncertainty, but when it is coupled which the changing configuration of the robot, it isn't easy to forge a structure for it. There are established methods of variable-structure controllers, such as adaptive controllers, to guarantee the performance of a system in the presence of bounded structured uncertainties \cite{ZHANG2017}. However, the performance of adaptive controllers is just guaranteed within the bounds of the structured uncertainty of the model of the system. Any unstructured uncertainties can throw the system's performance off its track. There are several attempts to exploit the benefits of both robust and adaptive controller schemes simultaneously, without giving up much of their benefits \cite{Li2008, Yao2010}. 

Various approaches have been proposed to design a controller for joints of a robotic manipulator \cite{Ferrara2015, YUAN2020, Garcia2020}. The Proportional–Integral–Derivative (PID) cascaded control is still the predominant approach in the industry due to its simplicity and effectiveness. A well-tuned PID controller can satisfy most of the design objectives of a mechanical system. However, in the case of the mobile manipulator, where the parameters of the dynamics of the system change regularly based on various factors such as the mass and shape of the load it holds, the configuration it takes, or the variation of the velocity of the mobile base, a PID controller with fixed parameters may not be able to deliver the desired performance at all condition. Moreover, the system's behavior, which is controlled with a set of constant parameters, will deteriorate as the robot's components get old and worn out. In \cite{Garpinger2014}, the authors study the trade-off between increasing the robustness of the performance and lowering the performance requirements. They introduced an elegant graphical tool that provides an insight into this trade-off. The method described in this paper is a direct adaptive algorithm, for which the adaptation mechanism is optimized based on a Bayesian optimization method.   

The nonlinear adaptive PID controller is one of the successful cases of adding adaptability to a classic control approach \cite{Seraji_1998, Armstrong2001, Sun2009, Yao2010, Garpinger2014}. The effectiveness of this controller stems from the fact that it formulates fundamental control requirements for transient and steady-state regimes into a simple gain scheduling scheme. Each gain of a PID controller adjusts the behavior of the closed-loop system in a different region of operation. Increasing the proportional gain decreases the rise time but increases the overshoot and might lead to undesirable oscillations. The gain of the differential term has almost the opposite effect. Although the integral term may increase the overshoot and push the system toward instability, it is often necessary to be used to eliminate the steady-state error. The nonlinear adaptive PID controller exploits the fact that the corresponding effects of each component of the PID can be scheduled based on the magnitude of the tracking error. When the output signal is far from its reference value, the proportional gain must be at its maximum to accelerate the system toward its goal. As the output comes near its reference value, the derivative gain needs to be increased to dampen any possible overshoot and trailing oscillations. 

We can implement the nonlinear adaptive PID control scheme with several choices for the nonlinear gain functions. However, almost all of these functions include hyper-parameters that must be chosen during the design process. These parameters determine how effective the gains of the PID controller will adjust to the changes in the system's state. The solution we propose in this paper to select these parameters is to examine various sets of parameters and indicate which one can optimize the performance measure we have chosen for our system during some extreme scenarios we have devised. Bayesian optimization provides a systematic approach to efficiently search for the optimal parameters \cite{Shahriari2016}. Initially, we may just use our expert view on choosing the hyper-parameters. These parameters should be chosen carefully such that the system stays stable. Unlike the gains of a PID controller, the relationship between the hyper-parameters and the performance of the system is more complicated. We may use a tuned linear PID controller as a clue toward the choice of minimum and maximum variation of a gain. After running the system with a handful of different sets of parameters, the information we acquire about the system may be used to train a model for the effect of changing the hyper-parameters on the performance of the system. Bayesian optimization uses this model to determine the next set of hyper-parameters to try on the real system, which may optimize its performance measure.


\section{Problem Statement}
\label{PROBLEM STATEMENT}

The objective of designing  controllers for a mobile manipulator is to move a payload from one point to another on a trajectory. The combined trajectory of the mobile base and the manipulator is devised in conjunction with each other; that is, the mobile base starts its motion as soon as the payload is securely grasped and while the manipulator is still moving the payload with respect to the base. This is in contrast with the cases in which the motion of the mobile base and the manipulator are devised separately, i.e., the mobile base moves to a point near the load, stops there until the robotic arm picks the load and secures it, and then the mobile base moves to its destination. 

The manipulator we study in this paper is equipped with an effort controller to actuate the joints. It draws feedback from the position of the joints and generates torque to move the robotic arm, such that the joint angles reach the desired values. As such, the robot's dynamics and load will directly affect the controller's performance. 

\subsection{Forward and Inverse Kinematics of a Mobile Manipulator}


\begin{figure}
       \centering
        \includegraphics[width=1\columnwidth]{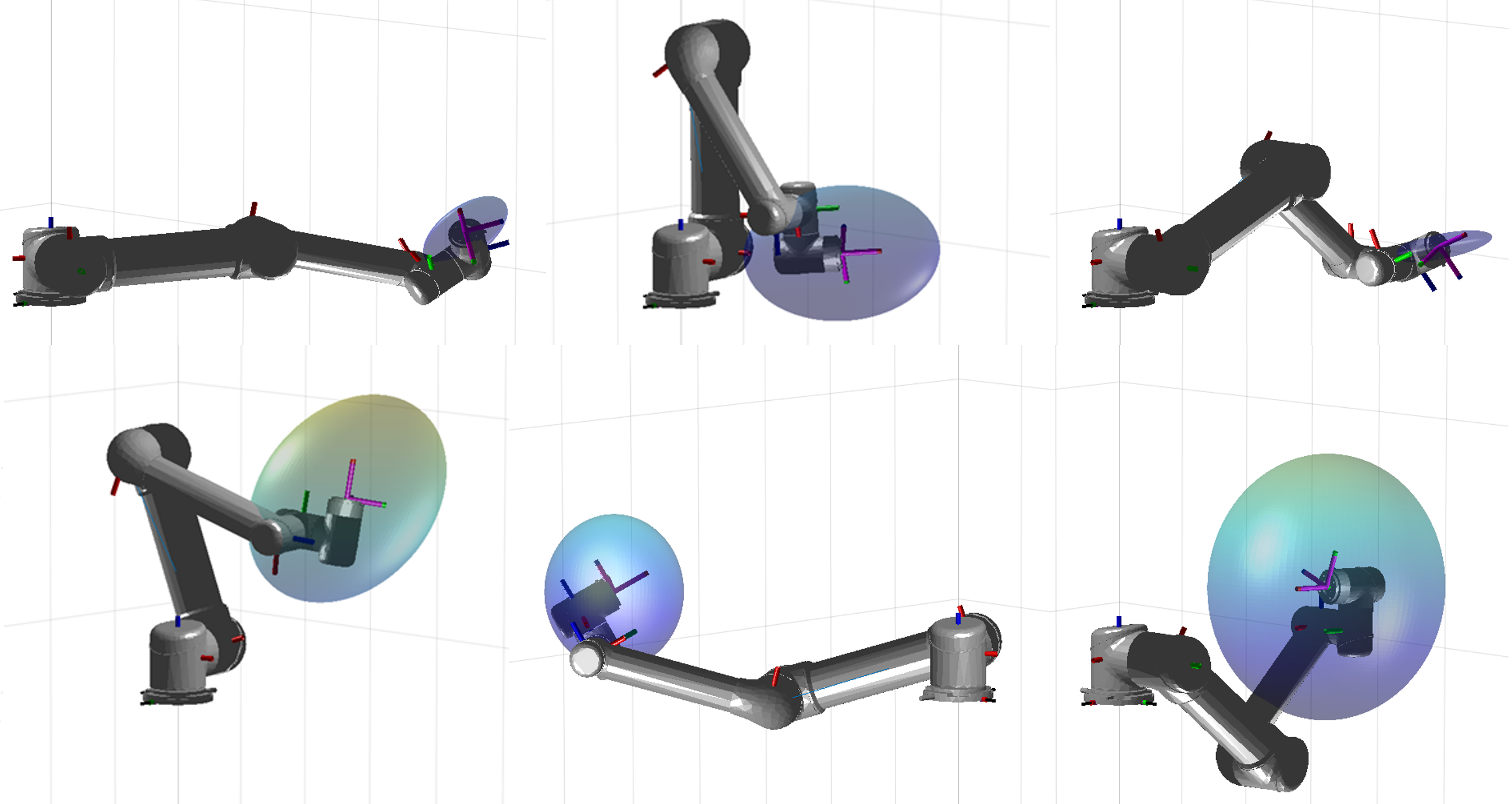}
        \caption{Compliance ellipsoid for UR5 in its different configurations. The compliance of the manipulator vary when the robotic arm is moving from one configuration to another; therefore, while there is a heavy load in hand and the base is moving, the effect of the Centripetal and Coriolis forces on the joint controllers change along the trajectory.}
        \label{Compliance depitcion}
             \vspace{-5mm}
\end{figure} 

The objective of designing controller for the joints of a manipulator is to have the joint variables follow the reference commands as closely as possible. In the case of a manipulator on a fixed base, this means compensating for the system's dynamics, including the dynamic of the manipulator and the effect of gravity and external forces on each joint. In the case of a mobile manipulator, additionally, the movement of the base generates centrifugal and Coriolis forces. These forces propagate back to the joints of the manipulator, and the actuator of the joint must overcome these forces to regulate the angle of the joint. Predicting the final load on the joints of a mobile manipulator is not trivial as they depend on the base's velocity and the arm's configuration at the time. Consider a manipulator with $n$ degrees of freedom. The dynamic equations of the system can be written as
\begin{equation}
    \mathbf{M}(\mathbf{q})\ddot{\mathbf{q}} +  \mathbf{C}(\mathbf{q},\dot{\mathbf{q}})\dot{\mathbf{q}} +  \mathbf{G}(\mathbf{q}) = \boldsymbol{T} + \mathbf{F},
    \label{Dynamic Equation}
\end{equation}
where $\mathbf{q}$ is the vector of the state variables, i.e., the joint variables, $\mathbf{M}(\mathbf{q}) \in \mathcal{R}^{n \times n}$ is the symmetric bounded positive definite inertia matrix, $\mathbf{C}(\mathbf{q},\dot{\mathbf{q}}) \in \mathcal{R}^{n \times n}$ represents the centripetal and Coriolis forces and torques, $\mathbf{G}(\mathbf{q}) \in \mathcal{R}^{n}$ delegates the overall effect of the the gravity on the manipulator, and $\boldsymbol{T}$ and $\mathbf{F}$ are the generalized forces and torques. 

In \eqnref{Dynamic Equation}, the centripetal and Coriolis forces,  $\mathbf{C}(\mathbf{q},\dot{\mathbf{q}})$, only encompass the effect of moving the manipulator. However, in the case of the mobile manipulator, it must be replaced with a more general term that also accounts for the motion of the base. The same is true for the first term, including the acceleration, which can be different in a non-inertial frame of the mobile base. The force generated at the end-effector of the robot, $\mathscr{F}$, and consequently the reactionary force, and the load observed by the joints, $\boldsymbol\tau$, are related by the Jacobian matrix $J$:
\begin{equation}
    \boldsymbol{\tau} = J^T \mathscr{F},
\end{equation}
where $\boldsymbol{\tau}$ are the torques generated with the actuators of the joints, and $\mathscr{F}$ is the generalized force, i.e., force and torque, which the end-effector exerts on its environment. In the case of a mobile manipulator, $\mathscr{F}$ should also compensate for the centripetal and Coriolis forces. We can also look at this equation from the point of view of the concept of compliance, i.e., if a force $\mathscr{F}$ is to be exerted on the end-effector, how much the end-effector will be displaced: 
\begin{equation}
    \delta \mathcal{X} = \pmb{\mathscr{C}} \mathscr{F}.
\end{equation}
The compliance matrix $\pmb{\mathscr{C}} = (J\mathbf{K}J^T)^{-1}$ is the function of the Jacobian matrix and is directly related to the configuration of the robot at any given time. $\mathbf{K}$ is the matrix of the stiffness of the joints. The eigen-structure of the compliance matrix, $\pmb{\mathscr{C}}$, provides us with information about the stability of the manipulator. \figref{Compliance depitcion} depicts the linear compliance of the Universal Robot (UR) against the force on its end-effector. The larger radius for the ellipse indicates the more compliant direction. The value of $\pmb{\mathscr{C}}$ changes according to the manipulator's configuration, and the motion of the base applies different forces according to its maneuver at any given time. These uncertainties erode the performance of any fixed structure controller. The controller of each joint should be able to adjust to these changes in order to deliver the predefined performance. In the rest of this paper, we will design and tune a nonlinear adaptive PID controller, which adjusts its parameter to deliver the predefined tracking performance in the presence of various uncertainties and disturbances mentioned above.

\section{Nonlinear Adaptive PID  Controller with Magnitude-Modulated Gains}
\label{Nonlinear Adaptive PID with magnitude-modulated gains}



PID controllers are commonly used as the controller of the actuators of industrial robots. A PID controller relates the control command to the error between the actual and the desired set-point, its integral, and derivative. Sequentially, the controller makes an effort to minimize the error, its integral over time, and the rate of change of the output. 

In a simple case of a stable minimum phase system, the proportional gain $K_{p}$ magnifies the error to push the output toward the set-point. Higher gain decreases the rise time for the system, which is desirable; however, when the system's output reaches the set-point and the error becomes zero---although only for a brief moment in time---the system will not stop abruptly. Even in the absence of any driving input, the inertia of the dynamic plant drives it beyond the set-point; the error increases again, although in a different direction; and the controller generates proportional effort to subdue this error too. The same happens this time in the opposite direction, which will continue until the error is smaller than a detectable boundary for the controller.

An oscillatory behavior for a joint is not desirable. It will erode the system's actuators, and at its extreme, it can destabilize the system. However, sometimes it is unavoidable due to the system's dynamics; this is often tolerated as the price to be paid for having an agile system. If we desire smoother output, we can introduce derivative terms as part of the PID controller. The derivative component produces control effort in proportion to the rate of change in the error and forces the system's output to settle with smaller ripples. But there is a price to pay for this smooth transition: the derivative term subdues the output changes not only around the set-point but also when it rises toward the set-point, i.e.,  the derivative term also increases the system's rise time.

The existence of the integral component sometimes is necessary to eliminate the steady-state error and improve the tracking performance of the system. However, it also comes with a cost: it amplifies the system's oscillation and pushes the system toward instability. Thus, the integral term is to fulfill a special requirement, zero steady-state error, to control a particular class of systems in which the control loop does not include any integrator. 

It appears that the necessary components of the PID controller act in contrast to each other. A desirable characteristic of a system is to respond rapidly to the changes in its input. When the system is controlled with a PID controller, this can be satisfied with high proportional and low derivative gains. Another desirable behavior of the system is damped oscillations of its output and, consequently its shorter settling time. This can be realized with high derivative and low proportional gains. Ostensibly, these desired performance characteristics of a closed-loop control system contradict each other.
As a solution to this conundrum, in \cite{Seraji_1998}, it was proposed to coordinate the gains to enact strongly when needed and to be weakened when their effect is against the desired performance. This is realized by replacing the constant gains of the PID controller with nonlinear functions of tracking error, so that
\begin{align}
    \mathscr{U}(t) = \mathrm{K}_p(e) \times e + \mathrm{K}_i \times \int_0^t{e(\tau)d\tau} + \mathrm{K}_d(e) \times \frac{de}{dt},
\end{align}
where $\mathscr{U}(t)$ is the control input and $e(t)$ is the process error. The proportional and derivative gains are functions of the tracking error, but the integral gain is chosen to be constant. The stability of the system is sensitive to the changes in the integral gain, and this sensitivity depends to the proportional and derivative gains of the controller. Therefore, any region we choose to search for better integral gain may have pockets inside where the parameters can destabilize the system. And, the importance of the stability of the system outweighs any benefits we might see from using a nonlinear integral gain. The proportional gain $\mathrm{K}_p$ and the derivative gain $\mathrm{K}_d$ are chosen as functions of the tracking error:
\begin{align}
\label{NonlinearGains_1}
    \mathrm{K}_p = &\mathrm{k}_{p,max} - \frac{2(\mathrm{k}_{p,max} - \mathrm{k}_{p,min})}{\exp(-\tau_pe) + \exp(\tau_pe)},\\
    \label{NonlinearGains_2}
	\mathrm{K}_d = &\mathrm{k}_{d,max}  \exp(-\tau_de^2).
\end{align}
The proportional gain may vary between $\mathrm{k}_{p,max}$ and $\mathrm{k}_{p,min}$. This interval does not include zero. The derivative gain, on the other hand, is small when the error is large, and as the output of the system gets closer to its set-point, the derivative gain increases to its maximum, $\mathrm{k}_{d,max}$, to prevent undesirable oscillations. Intuitively, both proportional and derivative gains are adjusted to stabilize the system. The most prominent methods to find the stability criteria for a system that is controlled with a nonlinear PID are Aizerman's conjecture and the Popov's criterion. These theorems divide a nonlinear system into two parts: a part that can be approximately modeled with a linear system, and the nonlinear residue that is the difference between the linear model and the actual nonlinear system. Using the Aizerman's conjecture and the Popov's criterion, we can set bounds that if a measure of this residual nonlinear part stays within these bounds, the stability of the linear portion implies the stability of the nonlinear system  \cite{krstic1995nonlinear}. These are very useful tools since we have a comprehensive toolbox to control linear systems, while most of the systems we are to control are nonlinear. In this paper, we aim to control a robotic manipulator, which itself is a nonlinear system, with a nonlinear PID controller. This is one of the best case studies to apply Aizerman's conjecture the Popov's criterion to study the stability of the system, see \cite{Seraji_1998, Armstrong2001, Ouyang2008, Sun2009}. We will use the results of such studies to set the bounds of the search area for the gains we have introduced in Equations \eqref{NonlinearGains_1} and \eqref{NonlinearGains_2}.    

\subsection{Parameter Selection for Nonlinear Adaptive PID}

To optimize the performance of a system, ideally, we need a model that relates the values of the parameters that should be tuned to a measure of the system's performance, \eg a cost function. If a mathematical model of such a relationship exists, we may employ one of several existing optimization techniques to find a set of parameters that optimizes the cost function. However, there are cases, including the problem we are addressing in this paper, where an explicit relationship cannot be formulated, so we need to resort to black-box optimization methods. Then, we need to study the behavior of the cost function as the parameters of the system change. If we incrementally change the parameters and evaluate the cost function at several points, we can build a numerical version of its model and choose the parameters of the system by studying this model. However, if the evaluation of the cost function is very expensive do, the optimization method needs to be sample-efficient (requiring only a few function evaluations), to find the best possible parameters that deliver the system's best performance. In this paper, every evaluation of the performance of the system requires driving the robotic platform. The number of robotic experiments is limited to a small number due to wear and tear of the hardware and the time it takes to conduct the experiment; therefore, we need an optimization method that can work with a small dataset to model the behavior of the system and use that model to optimize the performance of the system. 

The cost function $J$ we intend to minimize is a combination of the joint control effort and the tracking error between the state of that joint and the command it has received:
\begin{equation}
J = \sum_{k = 0}^{N} \{p(\theta^i_{k}-\theta^i_{ref})^2+ q  u_{k}^2\}.
\label{eqn:cost_function}
\end{equation}
Here, $\theta^i_{k}$ is the sample for the $i^{th}$ joint angle, $\theta^i_{ref}$ is the desired reference value for the $i^{th}$ joint angle, and $u^i_{k}$ is the control effort at time step $k$. The coefficients $p$ and $q$ adjust the trade-off between the response time of the closed-loop system and the amount of the control effort paid to regulate the output. The objective is to find a set of design parameters for the controller gains at Equations \eqref{NonlinearGains_1} and \eqref{NonlinearGains_2} to reduce this cost function to its minimum.

The effect of changing the parameters of the nonlinear controller on the performance of the closed-loop system is not obvious. Moreover, the nonlinearity of the system we aim to control and the presence of various deterministic and stochastic uncertainties and disturbances make it difficult to derive a mathematical model from describing such a relationship. Although it is difficult and time-consuming, we can run the system with a set of parameters and calculate the cost function in association with those parameters. We opt to present the relationship between the defined cost function and the parameters which should be tuned as $J=f(\mathrm{k}_{p,min}, \mathrm{k}_{p,max}, \tau_p, \mathrm{k}_{d,max}, \tau_d)$, where $f$ is an unknown function. We can repeat this procedure for a few iterations and collect samples by evaluating this function. The question is how we should explore the regions of the parameter space to find the optimal values.

The search space of the parameters is vast, and it is not feasible to draw many samples and evaluate the cost function for these samples in order to identify the relationship between the parameters and the system's performance. Therefore, we need to extract as much information about the function as possible from the limited number of samples we have. We propose to use Bayesian optimization to achieve this goal. Bayesian optimization is an approach to efficiently optimizing black-box objective functions that are expensive to evaluate. It builds a surrogate for the objective function. The surrogate model is typically cheap to query and easy-to-train model. A commonly used surrogate model for the objective function is a Gaussian process (GP) \cite{Rasmussen2005}. Each conducted experiment will add to the training dataset of the GP, so that over time the surrogate matches the true (but unknown) objective closely. The trained surrogate model is used by an acquisition function \cite{Brochu2010, Shahriari2016}, which tells us what parameter setting for the controller to try out next on the robotic system. 

We have five parameters to adjust for the controller of each joint of the robot. The search space for the optimal set of values for these parameters is large. To confine our search to possible values, we first tune a conventional PID controller of each joint with the Ziegler-Nichols method. Then we use the gains of the tuned PID controller to generate good guesses for the parameters of nonlinear adaptive PID. These initial values are used to train the GP model to be used as the surrogate by  Bayesian optimization. 
Formally, a GP can be specified by its mean and covariance functions. Given a set of input points ${X}=\left\{ {x}_{1},{x}_{2},\ldots,{x}_{N}\right\} $ and corresponding observations $ {Y}=\left\{ y_{1},y_{2},\ldots,y_{N}\right\} $, such that $ y_{i}=f({x}_{i})+\epsilon$, where $\epsilon\sim\mathcal{N}(0,\sigma_{\epsilon}^{2}$) denotes Gaussian noise with zero mean and $\sigma_{\epsilon}^{2}$ variance, the GP can be written as $f({x})\sim\mathcal{GP}\left(\ m({x}),k({x},{x}')\ \right)$, where, $m({x})$ is the mean function and $k({x},{x}')$ is the kernel function \cite{Rasmussen2005}. Thus, given the kernel, the data, the predictive mean $\bar{f}({x}^{*})$ and variance $\mathbb{V}({x}^{*})$ at a query point ${x}^{*}$ are%
\begin{align}
\bar{f}({x}^{*}) & = \mathbb{E}\left[f({x}^{*})\left|{X},{Y},{x}^{*}\right.\right]=k({X},{x}^{*})^{T}{\Sigma}{Y} 
\\
\mathbb{V}({x}^{*}) & = k({x}^{*},{x}^{*})-k(X, x^*)^T{\Sigma}k(X, x^*)\label{eq:GP_cov_and_mean}
\end{align}%
where ${\Sigma}=\left(k({X},{X})+\sigma_{\epsilon}^{2}{I}\right)^{-1}$, and $k({x},{x}') =\sigma_{\text{SE}}^2\exp\left( {-\tfrac{r^2}{2l^2}} \right)$ with $r=\left\Vert {x}-{x}'\right\Vert$ is the squared exponential kernel.

In Bayesian optimization, an acquisition function is a utility that is used to pick the next point in the parameter space to be explored. In this paper, we choose Expected Improvement (EI) as the acquisition function, which has been shown to be useful in many practical applications \cite{Snoek2012}. Given the surrogate model derived from $\left[X, Y \right ]$ and \eqref{eq:GP_cov_and_mean} with the squared exponential kernel, we select the next point by optimizing the acquisition function, EI, defined as
\begin{equation}
   \text{EI}({x})=\left(\bar{f}({x})-{y^{best}}\right)\Phi\left( \alpha \right)+  \sqrt{\mathbb{V}({x})}\phi\left( \alpha \right),
\label{eq:expected improvement}
\end{equation}%
where $ \alpha = \tfrac{\bar{f}({x})-{y^{best}}}{\sqrt{\mathbb{V}({x})}} $, $y^{best}$ is the best sample so far, $ \phi \left(\cdot\right)$ is the standard probability density function and $\Phi\left(\cdot\right)$ is the standard cumulative distribution function. The EI function leads us to explore the regions we know less about by looking for the highest difference between the current optimum and the rest of the function. In summary, Bayesian optimization is run iteratively where in each iteration, a query point is evaluated, the surrogate model is updated with the new data sample, and EI is run to select the next data point.  


We start with the robot at which the joints are controlled with the PID controller tuned with the Ziegler--Nichols tuning method. Then, we replace the controller for the lowest joint---the joint between the base and shoulder-pan, with a nonlinear adaptive PID controller and optimize its parameters while the other joints are still controlled with conventional PID controllers. Once we are satisfied with the performance of the controller of the joint between the base and shoulder-pan, we proceed to replace the controller of the next joint---the joint between shoulder-pan and upper-arm. This time, we use Bayesian optimization to find the best set of parameters for the controller of the joint between shoulder-pan and upper-arm. Lastly, we replace the controller of the joint between upper-arm and fore-arm with a nonlinear adaptive PID controller, and optimize the parameters of this controller while the other two joints are being controlled by the nonlinear adaptive PID controllers.

\section{Simulation Results}
\label{Simulation Results}

\begin{figure}[t]
\centering
    \includegraphics[width=1\linewidth]{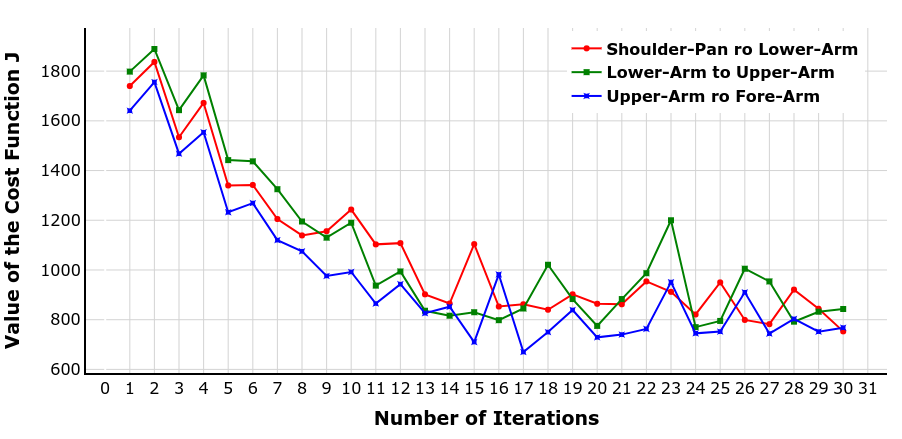}
    \caption{The value of the cost function (\eqnref{cost-function}) during the Bayesian optimization iterations. The coefficients of the cost function are $p = 1$ and $q = 0.5$, and $N = 246$.}
    \label{cost-function}
    \vspace{-15pt}
\end{figure}

\begin{figure*}
	\centering
	\rulesep
    \begin{subfigure}[b]{0.43\textwidth}
       \centering
       \includegraphics[width=1\textwidth]{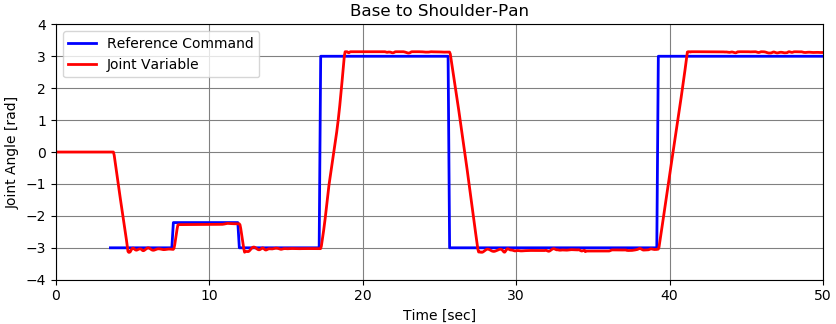}
    \end{subfigure}\rulesep
   \begin{subfigure}[b]{0.43\textwidth}
       \centering
       \includegraphics[width=1\textwidth]{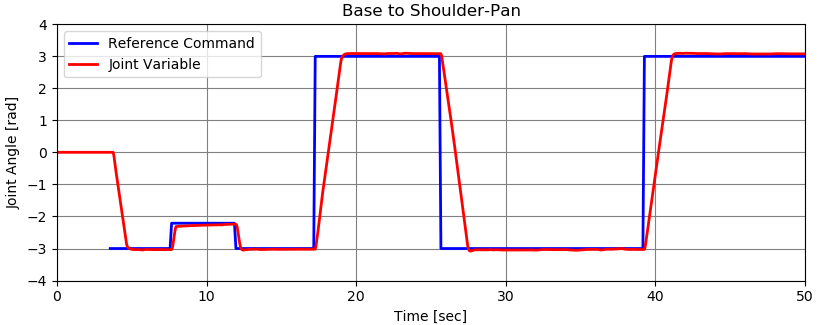}        
    \end{subfigure}\rulesep\\
    \rulesep
    \begin{subfigure}[b]{0.43\textwidth}
        \centering
        \includegraphics[width=1\textwidth]{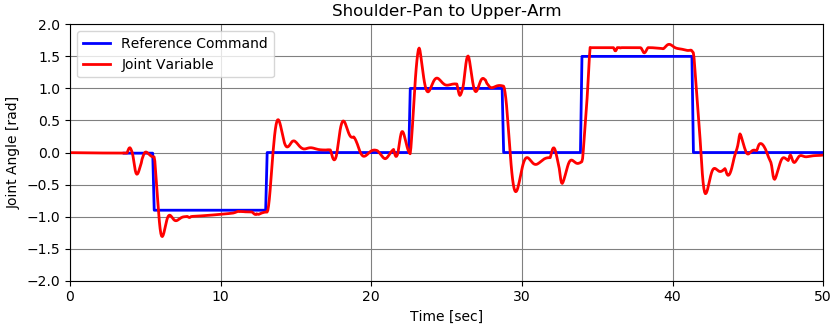}
    \end{subfigure}\rulesep
    \begin{subfigure}[b]{0.43\textwidth}
       \centering
       \includegraphics[width=1\textwidth]{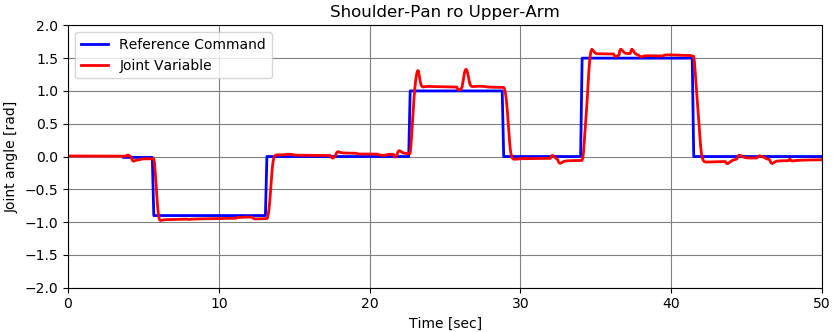}
    \end{subfigure}\rulesep\\
   \rulesep\begin{subfigure}[b]{0.43\textwidth}
       \centering
        \includegraphics[width=1\textwidth]{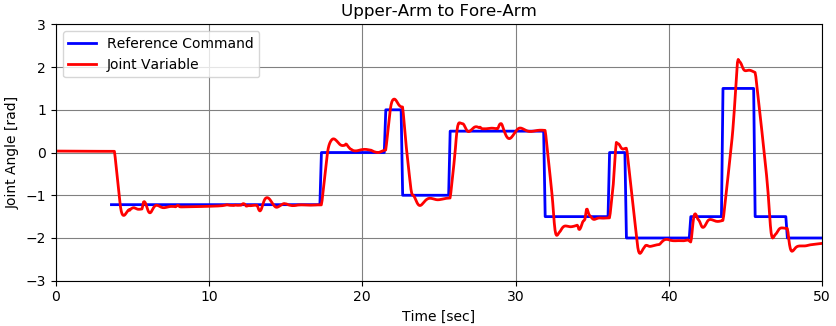}
        \caption{Performance of \textbf{Tuned PID}}
    \end{subfigure}\rulesep
    \begin{subfigure}[b]{0.43\textwidth}
        \centering
        \includegraphics[width=1\textwidth]{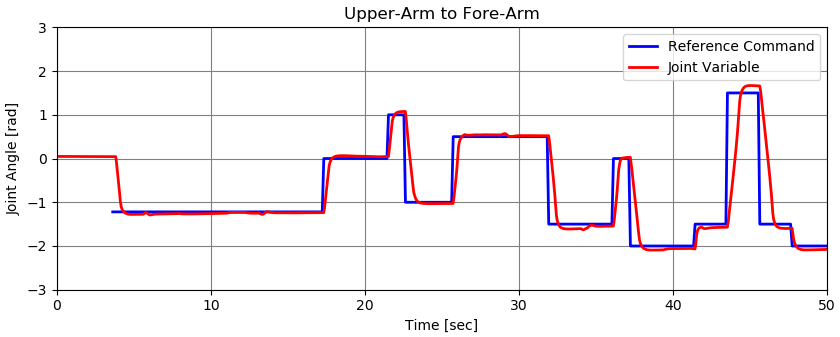}
        \caption{Performance of \textbf{Nonlinear Adaptive PID}}
    \end{subfigure}\rulesep
  \caption{The joint variables controlled with tuned PID controller (left), and nonlinear adaptive PID controller (right). The PID controller is tuned with the Ziegler--Nichols tuning method. Some fine-tuning is done manually afterward. The nonlinear adaptive PID controller is initialized with minimum and maximum values around the previously tuned values. Then, the trials are run based on the parameters Bayesian optimization suggests to fine-tune these minimum and maximum values and the time constants of the nonlinear gains. } 
  \label{fig: Control Performance}
\end{figure*}

\begin{figure*}
	\centering
	\rulesep
    \begin{subfigure}[b]{0.43\textwidth}
       \centering
       \includegraphics[width=1\textwidth]{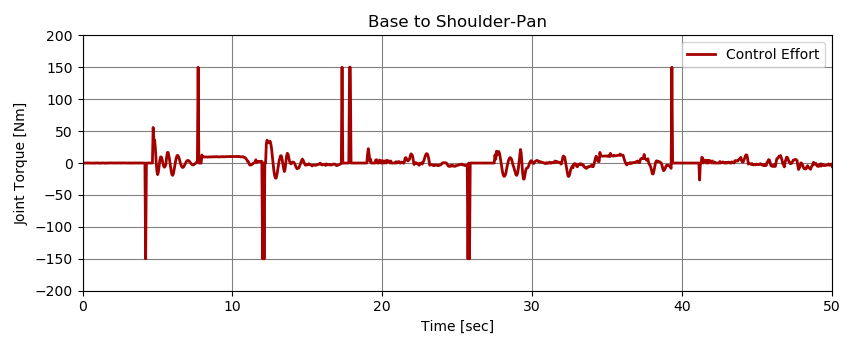}
    \end{subfigure}\rulesep
   \begin{subfigure}[b]{0.43\textwidth}
       \centering
       \includegraphics[width=1\textwidth]{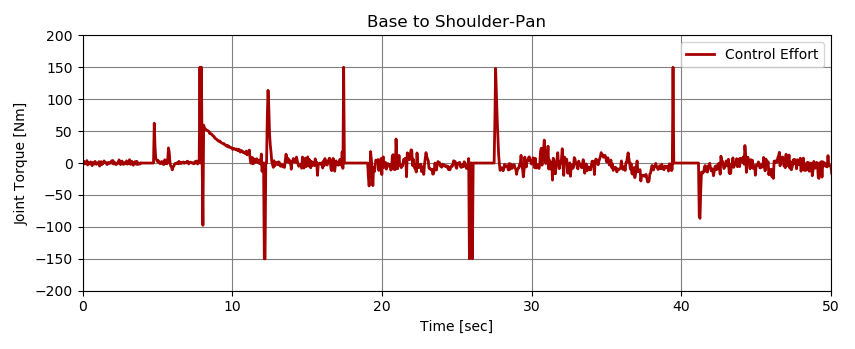}        
    \end{subfigure}\rulesep\\
    \rulesep\begin{subfigure}[b]{0.43\textwidth}
        \centering
        \includegraphics[width=1\textwidth]{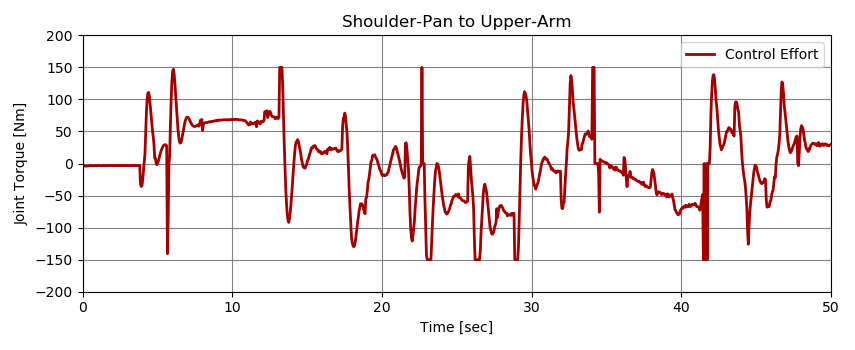}
    \end{subfigure}\rulesep
    \begin{subfigure}[b]{0.43\textwidth}
       \centering
       \includegraphics[width=1\textwidth]{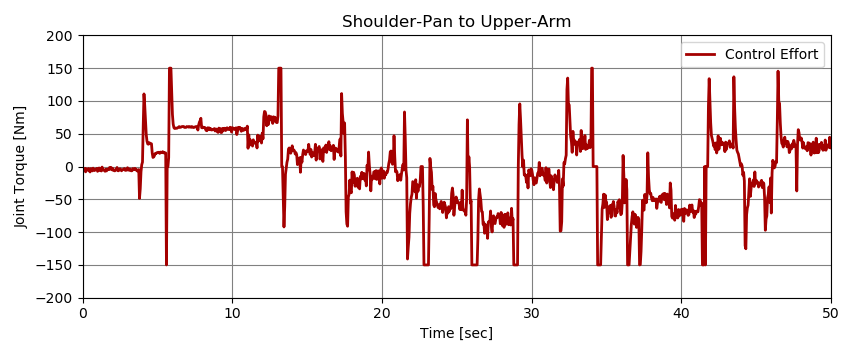}
    \end{subfigure}\rulesep\\
   \rulesep\begin{subfigure}[b]{0.43\textwidth}
       \centering
        \includegraphics[width=1\textwidth]{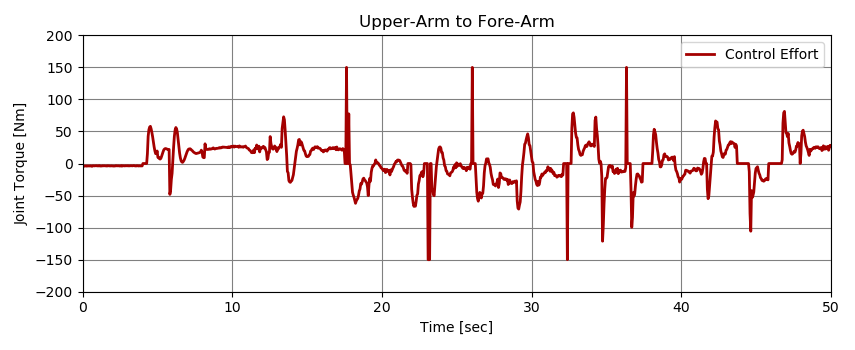}
        \caption{Control Efforts  - \textbf{Tuned PID}}
    \end{subfigure}\rulesep
    \begin{subfigure}[b]{0.43\textwidth}
        \centering
        \includegraphics[width=1\textwidth]{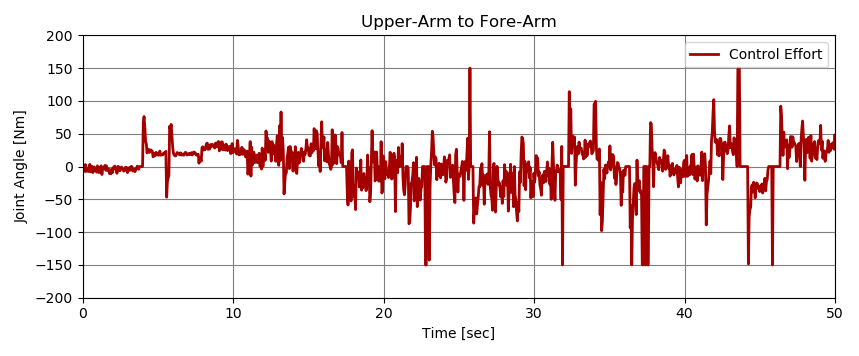}
        \caption{Control Efforts  - \textbf{Nonlinear Adaptive PID}}
    \end{subfigure}\rulesep
  \caption{The control effort of joints controlled with tuned PID controller (left), and nonlinear adaptive PID controller (right). The parameters of the nonlinear adaptive controller is chosen to decrease the tracking error and minimizes control effort.}
  \label{fig: Control Efforts}
\end{figure*}

\begin{figure*}
	\centering
	\rulesep
    \begin{subfigure}[b]{0.43\textwidth}
       \centering
       \includegraphics[width=1\textwidth]{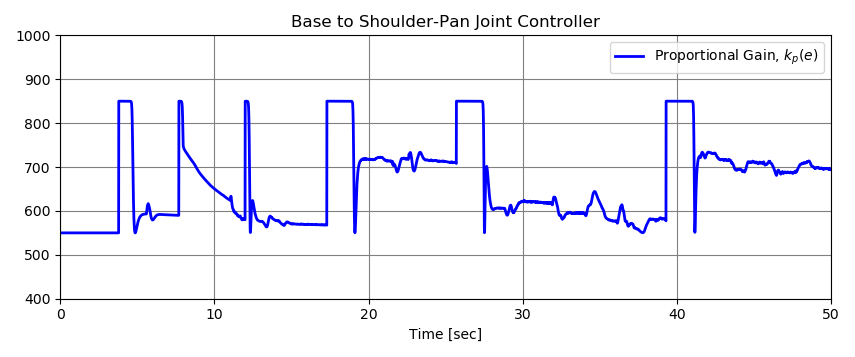}
    \end{subfigure}\rulesep
   \begin{subfigure}[b]{0.43\textwidth}
       \centering
       \includegraphics[width=1\textwidth]{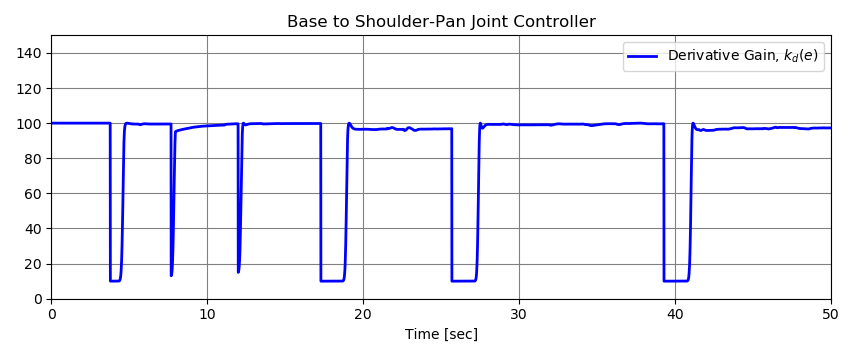}        
    \end{subfigure}\rulesep\\
    \rulesep\begin{subfigure}[b]{0.43\textwidth}
        \centering
        \includegraphics[width=1\textwidth]{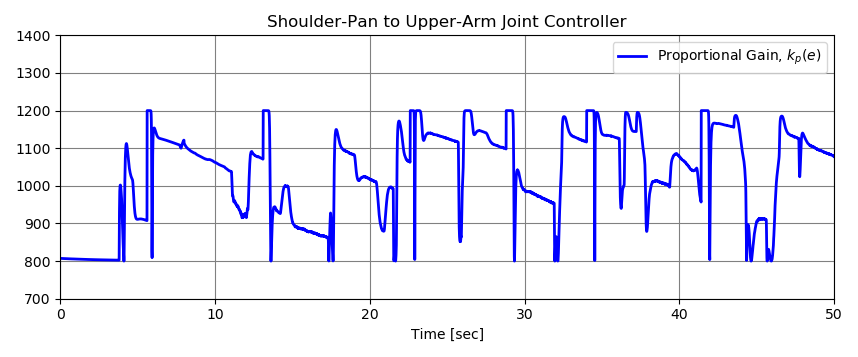}
    \end{subfigure}\rulesep
    \begin{subfigure}[b]{0.43\textwidth}
       \centering
       \includegraphics[width=1\textwidth]{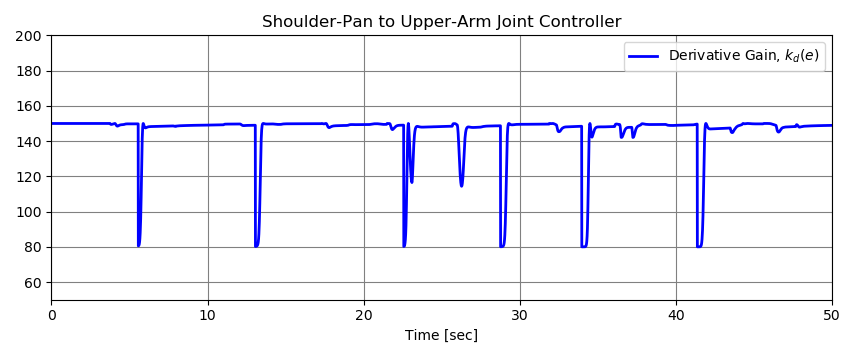}
    \end{subfigure}\rulesep\\
   \rulesep\begin{subfigure}[b]{0.43\textwidth}
       \centering
        \includegraphics[width=1\textwidth]{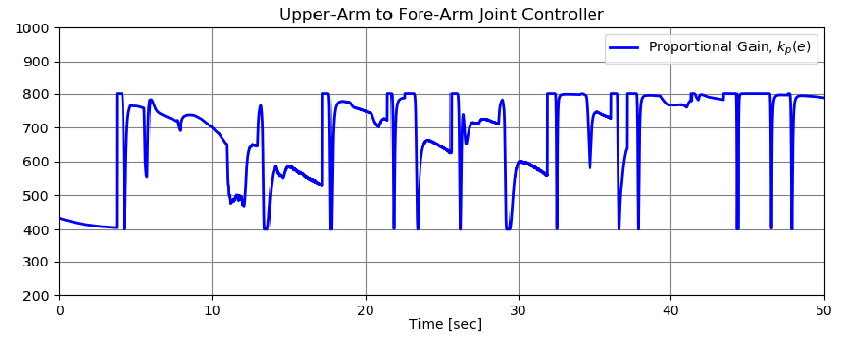}
        \caption{\textbf{Proportional Gain}}
    \end{subfigure}\rulesep
    \begin{subfigure}[b]{0.43\textwidth}
        \centering
        \includegraphics[width=1\textwidth]{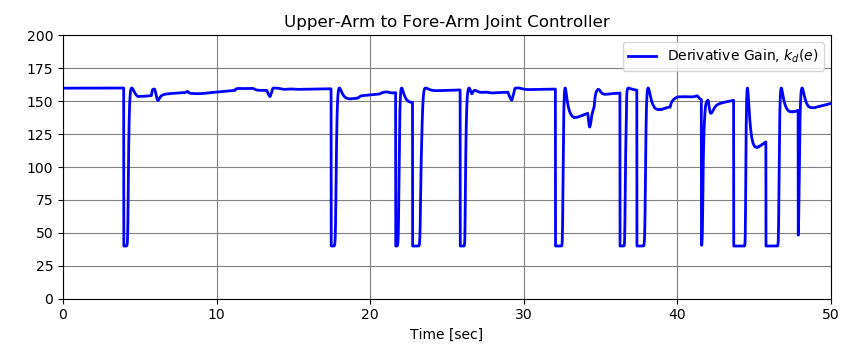}
        \caption{\textbf{Derivative Gain}}
    \end{subfigure}\rulesep
  \caption{Variation of the gains of the nonlinear adaptive PID controller of each joints. The proportional gain of the controller is responsive to disturbances. When the error between the output and the reference value is large the proportional gain is high to force the output back toward the reference value. The derivative gain is minimum at the beginning to let the joint move fast. As the tracking error decreases, the differential gain rises to its maximum and increases the stiffness of the joint.}
  \label{fig: Control Parameters}
  \vspace{-15pt}
\end{figure*}

We propose a controller to regulate the joint variables of a robotic arm of a mobile manipulator while it is transporting a heavy payload. The mobile manipulator we use is a MiR-100 mobile base from Mobile Industrial Robots\textsuperscript{\textregistered} on which a Universal Robot\textsuperscript{\textregistered} UR5 robotic arm is mounted. The end-effector attached to the manipulator is a Barrett\textsuperscript{TM} Hand model BH282. The load is moved along a convoluted trajectory, i.e., the mobile base starts moving while the arm brings in the load. \figref{simulation scenes} depicts a few scenes from the simulation runs to test the performance of the proposed controller. In this section, we present the results of the simulations of the proposed method. The simulation environment in which our robot operates is Gazebo, and we use the facilities of the Robot Operating System (ROS) to implement our solution. 

In the simulation, the mobile manipulator grasps a heavy object---a power drill (for which the weight is set to 3.5kg) and moves it around with purposefully designed convoluted and exaggerated motions. The motions include acceleration of the mobile base and sudden stops, turning sharply while the stretched-out arm is pulled in, and other maneuvers of alike, which are designed to exert a lot of load on the joints of the arm. Motions of the base and the manipulator are designed in conjunction with the motion of the arm to exert considerable centrifugal and Coriolis disturbance forces to the joints of the manipulator. This is to put the abilities of the controllers of the joints of the robot into a test to reject the possible external disturbances and regulate the joint variables under stress. In about 20 iterations of the Bayesian optimization loop, we see significant improvement, and we can find the good parameters for the controller of each joint, see \figref{cost-function}. Each trajectory in an iteration takes 25 seconds to simulate. After the 20th iteration is done, we choose the best set of parameters that yields the minimum value for the cost function. For upgrading the controllers of the joints, we start by replacing the controller of the lowest joint, i.e., the joint between the shoulder-pan and the lower-arm. Then, we continue to the next joint until the controllers of all three joints are replaced with the nonlinear adaptive PID controller.


\figref{fig: Control Performance} depicts the performance of the nonlinear adaptive PID alongside the tuned PID controller for each of three load-bearing joints of the UR5 manipulator. The nonlinear adaptive PID leads to less rise time and less overshooting. It damps disturbances more effectively. These are done at the cost of slightly higher control effort, see \figref{fig: Control Efforts}. The control effort for each joint of the manipulator is limited with an upper bound. In this simulation, the limit of the torque for a joint is set to $150.0Nm$. \figref{fig: Control Parameters} illustrates how the parameters of the controller change during the cycles of performance of the system. The derivative gain is less sensitive to disturbances. This behavior is desirable since this element of the PID controller is to keep the output of the system steady in the presence of external disturbances. The derivative gain is at its minimum when the joint receives the command. At this moment, the proportional gain is at its maximum to drive the output toward the desired reference command value. When the joint variable settles down to the desired value, the proportional gain is reduced, and the sensitivity of the output to the external disturbances is subdued. At this state, the derivative gain is high to dampen any output oscillation in response to the system's inertia or external disturbances.

\section{Conclusions}
\label{Conclusions}

In this paper, we design and optimize a nonlinear adaptive PID controller to regulate the joint variables of a mobile manipulator. The joints of a mobile manipulator must follow a command they receive despite any disturbances the motion of the mobile base can cause. The parameters of a PID controller balance a trade-off between different requirements which is needed to be satisfied during the different phases of the response of the system. In a mobile manipulator, where the the controller of the joints should endure large disturbances---because of the moving base and the load the manipulator is supposed to carry, the performance requirements have to be carefully balanced. We propose to use an adaptive scheme to adjust the gains of the PID controller according to the tracking error. When the reference value for a joint angle changes and the error between the current and the desired angle increases, the response of the system needs to be nimble and fast. Therefore, the proportional gain must be at its maximum. As the output comes closer to the reference value, the derivative gain must increase to damp down possible overshoot. The derivative gain stays high to help the system regulate the output in presence of any external disturbances. 

We used a Bayesian optimization approach to search for the best set of parameters for the adaptation mechanism of the gains for the proposed nonlinear adaptive PID controller. Our results demonstrate that the resulting control scheme improves the performance of the system in comparison to the baseline PID controller, which is tuned with the help of Ziegler--Nichols tuning method. In future work we examine how robust performance of the manipulator may improve the overall stability and performance of the mobile manipulator.

\bibliographystyle{IEEEtran}
\bibliography{references}

\end{document}